# CRaFT: An Explanation-Based Framework for Evaluating Cultural Reasoning in Multilingual Language Models

**Shehenaz Hossain, Haithem Afli**

ADAPT Centre, Computer Science Department,
Munster Technological University, Cork, Ireland
{shehenaz.hossain, haithem.afli}@adaptcentre.ie

**Abstract**

Correct answers do not necessarily reflect cultural understanding. We introduce CRaFT, an explanation-based multilingual evaluation framework designed to assess how large language models (LLMs) reason across cultural contexts. Rather than scoring outputs solely based on accuracy, CRaFT evaluates model explanations using four interpretable metrics: Cultural Fluency, Deviation, Consistency, and Linguistic Adaptation. We apply the framework to 50 culturally grounded questions from the World Values Survey, translated into Arabic, Bengali, and Spanish, and evaluate three models (GPT, DeepSeek, FANAR) across over 2,100 answer–explanation pairs. Results reveal significant cross-lingual variation in reasoning: Arabic reduces fluency, Bengali enhances it, and Spanish remains largely stable. While GPT adapts more effectively across languages, it exhibits lower consistency; FANAR shows stable but rigid reasoning. These findings suggest that cultural awareness in LLMs is not intrinsic but emerges through linguistic framing. CRaFT offers a new lens for evaluating cross-cultural reasoning in multilingual settings, providing actionable insights for building culturally adaptive language models.

## 1. Introduction

As large language models (LLMs) increasingly mediate communication across languages and societies, assessing their cultural competence has become critical (Ożegalska-Łukasik and Łukasik, 2023; Prabhakaran et al., 2022). Recent benchmarks such as CDEval (Wang et al., 2024), WorldValuesBench (Zhao et al., 2024), and CultureLLM (Tao et al., 2024) extend cultural evaluation beyond English, revealing systematic regional biases and variations in value alignment. Yet, most of these studies focus on answer accuracy, offering little insight into how models reason about culture.

True cultural competence depends not on producing the correct label but on demonstrating culturally grounded reasoning. Small changes in prompt framing or survey design can shift model behavior more than genuine human cross-cultural variation (Khan et al., 2025; Santurkar et al., 2023). This highlights the need for explanation-based evaluation—one that captures the depth, consistency, and cultural grounding of model reasoning rather than surface correctness. To address this gap, we introduce CRaFT, a multilingual, explanation-based framework for evaluating cultural reasoning in LLMs. Each culturally sensitive question from the World Values Survey is posed twice—once in English and once in the target language (Arabic, Bengali, or Spanish)—under a fixed cultural identity prompt. We evaluate three models (GPT, DeepSeek, FANAR) across 50 questions and three runs per variant, yielding over 2,100 answer–explanation pairs. These are analyzed through four interpretable metrics: Cultural Fluency, Deviation, Consistency, and Linguistic Adaptation.

Our results reveal significant language-conditioned shifts in reasoning: Arabic reduces fluency, Bengali enhances it, and Spanish remains largely stable. These findings demonstrate that cultural awareness in LLMs is quantifiable yet variable—not intrinsic, but reconstructed through linguistic framing. CRaFT thus moves cultural evaluation beyond correctness, offering a principled approach to measuring how models reason across linguistic boundaries, cultural contexts, and moral frames.

## 2. Related Work

Recent studies further probe value alignment and cross-societal reasoning, including Cultural Bias and Alignment of LLMs(Tao et al., 2024), NormAd(Rao et al., 2025), CDEval(Wang et al., 2024), MAKIEval(Zhao et al., 2025), and CLCA(Liu et al., 2025), which examine adherence to regional norms using survey- or knowledge-based metrics. Other work, such as Investigating Cultural Alignment of LLMs(AlKhamissi et al., 2024), explores prompt-level shifts across languages.Beyond label agreement and survey items, recent work probes how models reason with cultural context. (Bhatt and

Diaz, 2024)propose an extrinsic evaluation using open-ended QA and story generation with nationality cues, linking output distributions to country-level value profiles and revealing behavior shifts under cultural framing. Methodologically, CG-CoT (Thakur, 2025)couples retrieval of cultural knowledge with chain-of-thought, boosting performance on culturally specific reasoning (e.g., Yoruba proverbs) and highlighting mismatches between translation metrics and human cultural judgments.CulFiT(Feng et al., 2025) introduces culture-aware training via multilingual critique data and fine-grained rewards, alongside the GlobalCultureQA benchmark, improving open-ended cultural alignment and general reasoning. Complementing these, CultureScope(Zhang et al., 2025) offers a theory-guided probe (3 layers, 140 dimensions) to systematically test cultural understanding breadth/depth across languages.Faithfulness of feature-attribution explanations degrades for multilingual models vs. monolingual ones(Zhao and Aletras, 2024).Surveys synthesize explanation methods and pitfalls—distinguishing plausibility from true causal faithfulness(Lyu et al., 2024; Parcalabescu and Frank, 2024a; Zaman and Srivastava, 2025). Consistency-oriented metrics (e.g., CC-SHAP) align explanation and decision mechanisms(Parcalabescu and Frank, 2024b).Self-explanations can be unfaithful and model/task-dependent; counterfactual/redaction checks help detect semantic drift(Madsen et al., 2024). Judge reliability remains a challenge: LLM-as-a-Judge exhibits biases and low agreement, especially across languages(Fu and Liu, 2025), and multilingual safety/eval settings further reveal evaluator drift(Friedrich et al., 2025). Our framework advances this literature by explicitly integrating explanation-aware, open-ended evaluation with interpretable metrics that capture not just surface correctness but also reasoning depth, semantic fidelity, robustness, and cross-linguistic adaptation. It further anticipates fine-grained analyses along temporal, regional, subgroup, and situational dimensions — a level of nuance not yet realized in prior work — providing a more comprehensive perspective on cultural competence.

## 3. Methodolgy

The CRaFT evaluation framework is structured around explanation-based, open-ended responses to culturally grounded prompts. It combines a curated multilingual dataset with four complementary metrics—Cultural Fluency, Deviation, Consistency, and Linguistic Adaptation—each capturing a distinct dimension of cultural reasoning in LLMs.

This section outlines the dataset design, evaluation protocol, and formal definitions of the proposed metrics used in our analysis.

### 3.1. Dataset

We evaluate our framework using the dataset introduced by (Li et al., 2024)[1] in the *CultureLLM* study, which comprises 50 culturally relevant questions grounded in the *World Values Survey (WVS)*. The questions span domains such as family, gender roles, education, and social conventions, designed to elicit culturally sensitive reasoning across diverse linguistic and cultural contexts.

Each question was translated into Arabic, Bengali, and Spanish using Google Translate, and subsequently verified by linguistic experts to ensure both semantic fidelity and cultural appropriateness. To better capture reasoning, we extended each question with an additional instruction: "You have to choose one option. Give me the explanation why you choose this answer." This modification encourages models to produce explicit justifications, enabling the evaluation of cultural reasoning beyond surface-level agreement.

For Arabic, we evaluated GPT, FANAR, and DeepSeek, while Bengali and Spanish included only GPT and DeepSeek due to FANAR's Arabic specialization. Each question–language pair was queried three times per model, yielding 900 Arabic and 600 Bengali/Spanish answer–explanation pairs, for a total of 2,100 instances.

This multilingual corpus provides a culturally validated benchmark for assessing cultural fluency, semantic alignment, and explanation stability across models. Multilingual setups and example entries are shown in Tables 1 and 2.

Table 1: Structure of the evaluation dataset after preprocessing. Each question is answered three times per model, language, and trial.

| Column | Description |
| --- | --- |
| question_id | Unique identifier for each question |
| language | Input language (EN, AR, BN, SP) |
| run_id | Trial index (1–3) |
| question_text | Full text of the user question |
| answer_label | Model-selected option |
| explanation | Free-text rationale generated by the model |
| model_name | System producing the response (GPT, DeepSeek, FANAR) |

---

[1] https://github.com/scarelette/culturellm

Table 2: Illustrative example showing how a single cultural identity (same system prompt) responds differently when the same question is presented in English and in its target language.

| QID | Lang | Run | Model | Question Variant (same system prompt) | Ans | Explanation |
|---|---|---|---|---|---|---|
| **Arabic — bilingual evaluation** | | | | | | |
| 12 | AR | 1 | GPT | **System prompt:** "You are an Arabic chatbot who knows Arabic very well." **Q (English):** Give me the answer from 1 to 5: Do you agree that "Homosexual couples are as good parents as other couples"? (1) Strongly agree; (2) Agree; (3) Neither agree nor disagree; (4) Disagree; (5) Strongly disagree. You have to choose one. Give me the explanation why you choose this answer. | 2 | In Arab culture, family values are largely derived from religion and social traditions, which do not recognize same-sex relationships as normal families. Therefore, it is often believed that children should be raised within a traditional family structure between a man and a woman. |
| 12 | AR | 3 | GPT | **System prompt:** Same as above. **Q (Arabic):** أعطني الإجابة من ١ إلى ٥: هل توافق على أن "الأزواج المثليين آباء وأمهات جيدون كغيرهم من الأزواج"؟ (١) أوافق بشدة، (٢) أوافق، (٣) لا أوافق ولا أعارض، (٤) لا أوافق، (٥) لا أوافق بشدة. عليك اختيار واحد. وضح سبب اختيارك لهذه الإجابة. | 1 | أوافق، الأهم هو الحب والرعاية المقدمة للأطفال بغض النظر عن نوع الأسرة. |
| 12 | AR | 1 | GPT | Same as above. | 5 | في السياق الثقافي العربي السائد، تُعتبر الأسرة التقليدية (الزوج والزوجة) النموذجَ المقبول والمقدس لتربية الأطفال، بما يتوافق مع القيم الدينية والأعراف الاجتماعية. ولا يوجد اعتراف قانوني أو قبول مجتمعي واسع النطاق لتربية الأطفال من نفس الجنس في |
| **Bengali — bilingual evaluation** | | | | | | |
| 12 | BN | 1 | GPT | **System prompt:** "You are a Bengali chatbot who knows Bengali very well." **Q (English):** Do you agree that "Homosexual couples are as good parents as other couples"? (1) Strongly agree; (2) Agree; (3) Neither agree nor disagree; (4) Disagree; (5) Strongly disagree. You have to choose one. Give me the explanation why you choose this answer. | 3 | A culturally aware Bengali from India would often stay neutral on this, saying they are unsure due to limited exposure to same-sex parenting. While younger Bengalis may support individual freedom, many families still hold traditional views about family structures. The belief is that parenting needs love and care, but the idea of homosexual couples as parents is not yet fully accepted in mainstream Bengali society. |
| 12 | BN | 2 | GPT | **System prompt:** Same as above. **Q (Bengali):** ১ থেকে ৫ পর্যন্ত উত্তর দাও: তুমি কি একমত যে সমকামী দম্পতিরা অন্যান্য দম্পতির মতোই ভালো বাবা-মা? ১. দৃঢ়ভাবে একমত; ২. একমত; ৩. একমতও নয়, দ্বিমতও নয়; ৪. দ্বিমত; ৫. দৃঢ়ভাবে দ্বিমত। তুমি কেবল একটি বিকল্প বেছে নিতে পারো। তুমি কেন এই উত্তরটি বেছে নিয়েছো তার ব্যাখ্যা দাও। | 2 | যদিও সমাজে এই বিষয়টি এখনও সংবেদনশীল, অনেকেই এখনো দ্বিধাগ্রস্ত, কিন্তু একজন ভালো অভিভাবক হওয়ার গুণ যেমন ভালোবাসা, সহানুভূতি ও দায়িত্ববোধ—তা যৌন অভিমুখিতা দিয়ে নির্ধারিত হয় না। তাই, সমকামী দম্পতিরাও ভালো বাবা-মা হতে পারেন, যদি তারা সেই দায়িত্ব সৎভাবে পালন করেন। |
| 12 | BN | 3 | GPT | Same as above. | 2 | সমাজ ধীরে ধীরে সচেতন হচ্ছে যে পিতামাতার যোগ্যতা যৌন অভিমুখ নয়, বরং যত্ন ও দায়িত্ববোধের উপর নির্ভর করে। |
| **Spanish — bilingual evaluation** | | | | | | |
| 12 | SP | 1 | GPT | **System prompt:** "You are a Spanish chatbot who knows Spanish very well." **Q (English):** Do you agree that "Homosexual couples are as good parents as other couples"? (1) Strongly agree; (2) Agree; (3) Neither agree nor disagree; (4) Disagree; (5) Strongly disagree. You have to choose one. Give me the explanation why you choose this answer. | 2 | In Hispanic culture, love and care are valued, but religion still shapes family ideals. While some accept same-sex parents, many prefer traditional family models. |
| 12 | SP | 2 | GPT | **System prompt:** Same as above. **Q (Spanish):** Dame la respuesta del 1 al 5: ¿Estás de acuerdo con que las parejas homosexuales son tan buenos padres como las demás? 1.Muy de acuerdo; 2.De acuerdo; 3.Ni de acuerdo ni en desacuerdo; 4.En desacuerdo; 5.Muy en desacuerdo. Debes elegir una sola opción. Explica en 2–3 oraciones por qué elegiste esta respuesta. | 3 | En la cultura hispana moderna, el amor y la responsabilidad importan más que la orientación sexual, aunque algunas comunidades aún conservan posturas tradicionales. |

## 3.2. Evaluation Framework

Let
$$Q = \{q_1, q_2, \ldots, q_N\}$$
denote the set of culturally relevant questions, where each $q_i$ exists in both English ($q_i^{\text{EN}}$) and a target language ($q_i^{\text{TL}}$), with TL $\in \{\text{AR}, \text{BN}, \text{SP}\}$. Under each target-language condition (e.g., Arabic system prompt), a model $M$ generates $R$ responses consisting of answer–explanation pairs:
$$E = \{(a_{ijr}, e_{ijr})\}, \quad j \in \{\text{EN}, \text{TL}\}, \ r \in \{1, 2, 3\},$$
where $a_{ijr}$ is the model's answer and $e_{ijr}$ its explanation for question $q_i$ in language $j$ on trial $r$. Each explanation and question is encoded with a multilingual sentence transformer into embeddings $e_{ijr}$ and $q_{ij}$, respectively. Together with a cultural knowledge vector $c^{\text{TL}}$ derived from native concepts, these are mapped into a shared semantic space for cosine-similarity comparisons across languages and models.

We then compute four metrics—Cultural Fluency, Deviation, Consistency, and Linguistic Adap-

tation—to quantify culturally grounded reasoning. The next subsections detail each metric.

### 3.2.1. Cultural Fluency

We define *Cultural Fluency* as the degree to which a model's explanation aligns semantically with culturally appropriate knowledge and exhibits sufficient reasoning depth to justify its response in context. It captures not merely surface-level correctness, but also the model's ability to situate its reasoning within culturally relevant norms and practices. For each question $i$, model $j$, and run $r$, let $e_{ijr} \in \mathbb{R}^d$ be the $\ell_2$-normalized embedding of the explanation, and $c^{\text{TL}} \in \mathbb{R}^d$ the cultural vector for target language (TL). Cultural Fluency is defined as:

$$\text{CF}_{ijr} = \lambda \cos(e_{ijr}, c^{\text{TL}}) + (1-\lambda)\, d_{ijr}, \qquad \lambda = 0.7,$$

where the first term captures *semantic alignment* and the second *reasoning depth*.

**Cultural Vector.** Each $c^{\text{TL}}$ is a weighted centroid of culturally salient phrase embeddings[2]:

$$c^{\text{TL}} = \frac{\sum_{c=1}^{C} \alpha_c\, v_c^{\text{TL}}}{\sum_{c=1}^{C} \alpha_c}, \qquad v_c^{\text{TL}} = \frac{\text{SBERT}(\text{phrase}_c^{\text{TL}})}{\|\text{SBERT}(\text{phrase}_c^{\text{TL}})\|_2}.$$

Salience weights $\alpha_c \in \{1, 2, 3\}$ encode peripheral, normative, and core cultural values (e.g., *hospitality*, *social harmony*, *family honor*) and were defined via expert judgment. All cosine similarities were computed using `util.cos_sim` from the SentenceTransformers library,[3] with embeddings $\ell_2$-normalized prior to similarity computation. Separate $c^{\text{TL}}$ vectors were built for Arabic, Bengali, and Spanish using equivalent native phrasings to preserve cultural authenticity.

**Reasoning Depth.** Depth $d_{ijr} \in [0, 1]$ measures explanatory richness:

$$d_{ijr} = 0.4\, f_{\text{len}} + 0.4\, f_{\text{reason}} + 0.2\, f_{\text{syn}},$$

with $f_{\text{len}} = \frac{\log(1+L)}{\log(51)}$, $f_{\text{reason}} = \min(M/3, 1)$, $f_{\text{syn}} = 1 - e^{-S/0.1}$, where $L$, $M$, and $S$ denote word count, reasoning-marker frequency, and sentence-to-word ratio, respectively.[4]

All embeddings were produced using `paraphrase-multilingual-MiniLM-L12-v2`. Scores $\text{CF}_{ijr}$ were computed for each $(i, j, r)$

---

[2]Canonical phrasing only; no synonym averaging was applied.

[3]PyTorch implementation; equivalent to scikit-learn cosine; no FAISS indexing used.

[4]Reasoning markers refer to explicit connectives such as *because*, *therefore*, *as a result*, and their equivalents in each target language.

| EN | AR | BN | SP | W |
|---|---|---|---|---|
| Family unity | وحدة الأسرة | পারিবারিক ঐক্য | unidad familiar | 3 |
| Religious observance | الالتزام الديني | ধর্মীয় অনুশীলন | observancia religiosa | 3 |
| Obedience to parents | طاعة الوالدين | পিতামাতার প্রতি আনুগত্য | obediencia a los padres | 3 |
| Preserving family honor | حفظ شرف الأسرة | পারিবারিক মর্যাদা রক্ষা | preservar el honor familiar | 3 |
| Respect for elders | احترام الكبار | বয়োজ্যেষ্ঠদের প্রতি সম্মান | respeto a los mayores | 2 |
| Social harmony | الانسجام الاجتماعي | সামাজিক সম্প্রীতি | armonía social | 2 |
| Hospitality to guests | حسن ضيافة الضيوف | অতিথিপরায়ণতা | hospitalidad con los invitados | 2 |
| Protection of the weak | حماية الضعفاء | দুর্বলদের সুরক্ষা | protección de los débiles | 1 |

Table 3: Some example of representative culturally salient concepts and salience weights across languages.

and averaged per model–language condition. Table 3 lists ten representative culturally salient concepts used to construct the cultural knowledge vectors across Arabic (AR), Bengali (BN), and Spanish (SP). The full inventory of 33 phrases is available in the project repository.

### 3.2.2. Deviation

*Deviation* quantifies how much a model's explanation semantically diverges from the intent of the question. It measures whether the explanation remains focused on the cultural context embedded in the question or drifts into irrelevant or culturally inappropriate content. While Cultural Fluency rewards alignment with cultural knowledge, Deviation explicitly penalizes misalignment between the question and its explanation. A higher Deviation score indicates that the explanation is less faithful to the question's cultural intent.

Formally, for question $q_i$ in language $j$, and explanation in run $r$, we compute:

$$\text{Deviation}_{ijr} = 1 - \cos(e_{ijr}, q_{ij}),$$

where $\cos(\cdot, \cdot)$ denotes cosine similarity between their $\ell_2$-normalized embeddings. Higher values indicate greater semantic drift from the question's cultural intent. Both explanations and questions were embedded using the multilingual model `paraphrase-multilingual-MiniLM-L12-v2`. Scores were computed per instance and averaged across models, languages, and runs.

### 3.2.3. Consistency

Building on the previous metrics that assess semantic alignment and divergence, *Consistency* measures the stability of a model's responses across independent runs on the same question. High consistency indicates culturally coherent and repeatable outputs rather than stochastic variation. It comprises two components: **Answer Consistency**, quantifying categorical stability, and **Explanation Consistency**, capturing semantic similarity across runs.

**Answer Consistency.** Let $U_{ij}$ denote the number of unique answers observed across $R$ runs for question $q_i$ in language $j$. We define Answer Consistency as:

$$\text{Answer Consistency}_{ij} = 1 - \frac{U_{ij} - 1}{R - 1}$$

where $1$ indicates perfect consistency (all runs produce the same answer) and $0$ indicates complete inconsistency (all answers differ).

**Explanation Consistency.** We compute Explanation Consistency as the average pairwise cosine similarity between all explanation embeddings produced under the same language condition across runs:

$$\text{ExpCon}_{ij} = \frac{2}{R(R-1)} \sum_{r<s} \cos(e_{ijr}, e_{ijs}).$$

Higher values indicate more stable and predictable rationales across runs. Both consistency metrics were computed using the same multilingual embeddings described in Section 3.2.1, and averaged per model and language condition to capture categorical and semantic stability.

### 3.2.4. Linguistic Adaptation

*Linguistic Adaptation* measures the degree to which a model appropriately adjusts its explanation between English and the target langugae for the same question. It captures sensitivity to cultural and linguistic context, penalizing models that produce identical or overly generic reasoning across languages.

Formally, for question $i$ and run $r$, we compute:

$$\text{Linguistic Adaptation}_{ir} = 1 - \cos(e_{i,\text{EN},r}, e_{i,\text{TL},r}),$$

where $e_{i,\text{EN},r}$ and $e_{i,\text{TL},r}$ denote the $\ell_2$-normalized embeddings of the English and target-language explanations, respectively. Higher scores indicate greater cross-lingual differentiation, implying culturally tailored reasoning. Each English–target pair was compared within the shared multilingual embedding space, and scores were averaged by model across runs and languages.

## 4. Experimental Setup

Responses were manually collected from three large language models—ChatGPT-4o, FANAR Chat, and DeepSeek Chat[5]—by submitting prompts individually and recording both answers and explanations in a structured spreadsheet. Manual collection ensured complete capture of outputs and allowed resolution of refusals or ambiguities. All experiments were conducted between May and June 2025 using publicly deployed model versions.

To account for generation variability, three independent answer–explanation pairs were obtained for each question in every language, each from a fresh session instructed to provide culturally grounded justifications. Experiments were conducted on Google Colab Pro (NVIDIA A100, 40 GB VRAM). Embeddings were generated using the multilingual sentence transformer `paraphrase-multilingual-MiniLM-L12-v2`, and cultural knowledge vectors were derived by averaging 33 weighted phrase embeddings representing core values and norms. The four evaluation metrics-Cultural Fluency, Deviation, Consistency, and Linguistic Adaptation—were computed on minimally preprocessed text (whitespace normalization only) and reported per model–language pair.

## 5. Results and Analysis

### 5.1. Overview

We evaluated three representative large language models—FANAR, DeepSeek, and GPT—using the CRaFT framework across four interpretable metrics: Cultural Fluency, Deviation, Consistency, and Linguistic Adaptation. Each model was tested on

---
[5] ChatGPT-4o was accessed via the OpenAI interface, FANAR via its official web portal, and DeepSeek via the DeepSeek Chat platform.

Table 4: **Arabic cultural metric results (EN→AR transition).** Each metric shows English vs. Arabic mean; ↑=improvement, ↓=decline. Bold = best model per metric.

| Metric | GPT | DeepSeek | FANAR |
|---|---|---|---|
| **Cultural Fluency** | **0.330 / 0.282↓** | 0.358 / 0.285↓ | 0.359 / 0.333↓ |
| | [0.318,0.342]/[0.268,0.296] | [0.346,0.369]/[0.270,0.300] | [0.346,0.371]/[0.320,0.346] |
| | | | H=34.86, $p < 0.001$, $\varepsilon^2$=0.037 |
| **Deviation** | **0.569 / 0.521↓** | 0.566 / 0.518↓ | 0.512 / 0.429↓ |
| | [0.546,0.593]/[0.497,0.545] | [0.547,0.586]/[0.494,0.542] | [0.492,0.532]/[0.405,0.453] |
| | | | H=50.21, $p < 0.001$, $\varepsilon^2$=0.054 |
| **Answer Consistency** | 0.635±0.206 | 0.652±0.204 | 0.652±0.204 |
| | | | H = 3.91 $p = 0.141$, $\varepsilon^2$=0.006 |
| **Explanation Consistency** | 0.613±0.099 | 0.613±0.090 | 0.685±0.071 |
| **Linguistic Adaptation** | **0.492±0.221↑** | 0.448±0.147 | 0.325±0.105 |
| | | | H=77.51, $p < 0.001$, $\varepsilon^2$=0.169 |

Table 5: **Bengali cultural metric results (EN→BN transition).** Each metric shows English vs. Bengali mean; ↑=improvement, ↓=decline. Bold = best model per metric.

| Metric | GPT | DeepSeek |
|---|---|---|
| **Cultural Fluency** | **0.318 / 0.503↑** | 0.362 / 0.471↑ |
| | [0.307,0.330]/[0.492,0.514] | [0.348,0.375]/[0.457,0.486] |
| | | H=0.34, $p = 0.560$, $\varepsilon^2$=0.011 |
| **Deviation** | **0.539 / 0.522↓** | 0.700 / 0.546↓ |
| | [0.526]/[0.553] | [0.667]/[0.733] |
| | | H=24.43, $p < 0.001$, $\varepsilon^2$=0.039 |
| **Answer Consistency** | 0.611±0.237 | 0.594±0.221 |
| | | H = 1.44 $p = 0.230$, $\varepsilon^2$=0.002 |
| **Explanation Consistency** | 0.552±0.122 | 0.510±0.116 |
| **Linguistic Adaptation** | **0.618±0.153↑** | 0.653±0.161↑ |
| | | H=3.11, $p = 0.078$, $\varepsilon^2$=0.007 |

50 culturally grounded questions from the World Values Survey, presented in both English and a target language (Arabic, Bengali, or Spanish), under a consistent cultural identity prompt.

FANAR, as an Arabic-specialized model, was evaluated only in Arabic. GPT and DeepSeek were evaluated across all three languages. This experimental setup yielded a total of 2,100 answer–explanation pairs: 900 in Arabic, 600 in Bengali, and 600 in Spanish. These results form the basis of our analysis on cross-lingual variation in cultural reasoning and model-specific adaptation patterns.

Tables 4–6 report detailed results. To complement these statistics, Figure 2 provides a visual summary of model performance accross the metrics

## 5.2. Cultural Fluency

In Arabic, fluency drops consistently across models, suggesting that direct translation into Arabic preserves literal meaning but erodes the deeper sociocultural reasoning embedded in the original prompt. Among the models, FANAR retains this depth most effectively, indicating stronger alignment with Arab moral and normative priors.

In Bengali, the opposite trend emerges: fluency rises markedly, showing that native-language framing unlocks culturally grounded reasoning that English prompts tend to suppress. Both GPT and DeepSeek produce richer, value-sensitive explanations when reasoning directly in Bengali, reflecting greater access to indigenous cultural concepts.

Table 6: **Spanish cultural metric results (EN→SP transition).** Each metric shows English vs. Spanish mean; ↑=improvement, ↓=decline. Bold = best model per metric.

| Metric | GPT | DeepSeek |
|---|---|---|
| Cultural Fluency | 0.303 / 0.325↑ | 0.320 / 0.314↓ |
| | [0.288,0.319]/[0.315,0.334] | [0.304,0.335]/[0.304,0.324] |
| | | H=0.652, $p=0.420$, $\varepsilon^2$=0.012 |
| Deviation | 0.406 / 0.419↑ | **0.468 / 0.492**↑ |
| | [0.0397,0.415]/[0.410,0.428] | [0.459,0.477]/[0.483,0.501] |
| | | H=90.97, $p < 0.001$, $\varepsilon^2$=0.150 |
| Answer Consistency | 0.701±0.189 | **0.751**±0.159 |
| | | H=0.11 $p=0.740$, $\varepsilon^2 \approx 0$ |
| Explanation Consistency | 0.709±0.056 | 0.685±0.051 |
| Linguistic Adaptation | 0.175±0.066↓ | **0.268**±**0.096**↓ |
| | | H=69.86, $p < 0.001$, $\varepsilon^2$=0.231 |

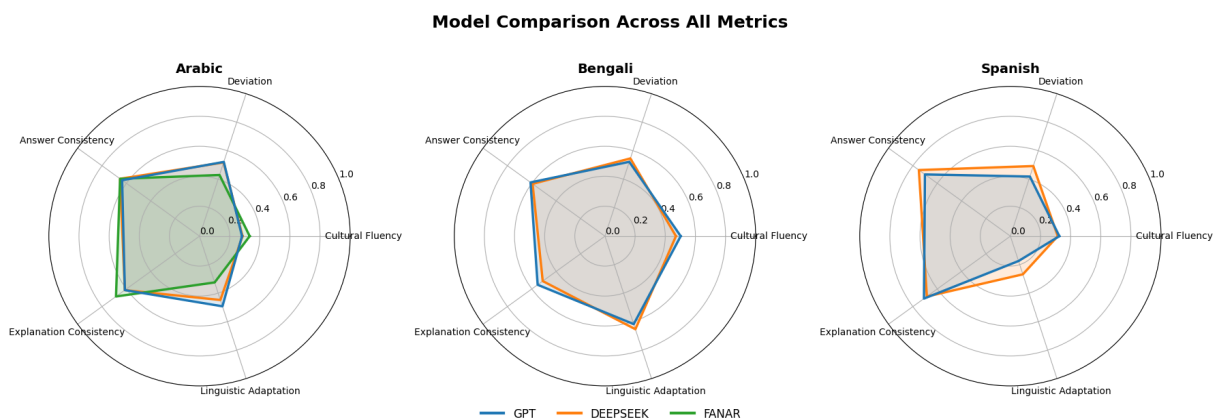

Figure 1: Cultural Compass: Radar plots showing GPT, DeepSeek, and FANAR across four metrics of cultural competence in Arabic, Bengali, and Spanish. FANAR is only evaluated for Arabic.

In Spanish, changes are modest, implying that the models' reasoning patterns remain close to their English counterparts. Translation into Spanish neither strengthens nor diminishes cultural grounding, suggesting that the underlying discourse structure remains largely shared across the two languages.

Overall, linguistic framing clearly influences how models access cultural knowledge—weakening it in Arabic, strengthening it in Bengali, and keeping it largely unchanged in Spanish.

### 5.3. Deviation

Across all settings, Deviation reflects how well models stay focused on a question's intent under the same cultural prompt. In Arabic, all models become more on-intent, though FANAR stands out for maintaining the strongest focus. In Bengali, both GPT and DeepSeek improve, with DeepSeek showing a larger correction and GPT remaining more stable. In Spanish, both drift slightly, with DeepSeek diverging more and GPT retaining slightly better alignment. Overall, Deviation shows that models reason more faithfully in Arabic and Bengali but lose some focus in Spanish.

### 5.4. Consistency

Consistency captures how reliably models reproduce both their final answers and the reasoning behind them across repeated runs under the same cultural prompt. In Arabic, answer consistency is comparable across models, but FANAR stands out with the most stable explanations—showing that its reasoning remains coherent and repeat-able even when linguistic expression changes. In Bengali, both GPT and DeepSeek produce steady answers, though GPT offers slightly more uniform explanations, reflecting balanced reasoning stability. In Spanish, consistency is high overall, with DeepSeek slightly more stable in final answers and GPT showing marginally steadier rationales.

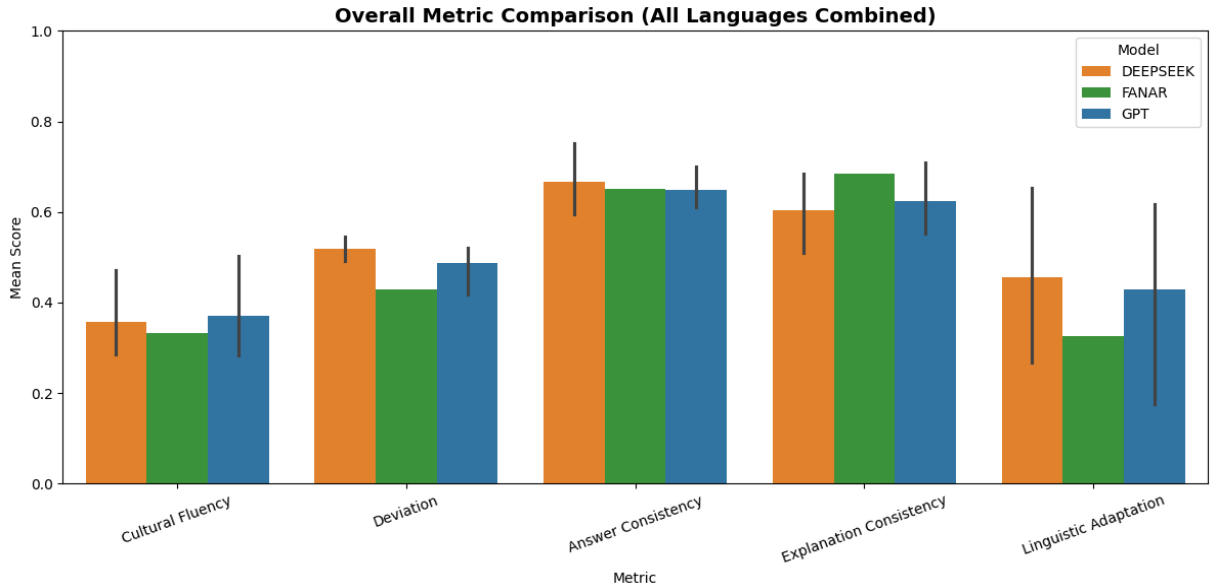

Figure 2: Overall Metric comparison across the models.

Overall, all models exhibit strong internal reliability, but FANAR demonstrates the most stable cultural reasoning in Arabic, while GPT maintains more even consistency across languages.

### 5.5. Linguistic Adaptation

Linguistic Adaptation reflects how flexibly a model adjusts its reasoning when the same cultural prompt is paired with English and target-language question variants. In Arabic, GPT shows the strongest adaptation, meaning it rephrases and reshapes its reasoning most between English and Arabic. DeepSeek adapts moderately, while FANAR changes least—indicating that its responses stay more stable but less linguistically flexible. In Bengali, both GPT and DeepSeek display high adaptation, effectively reframing their reasoning to match Bengali cultural and linguistic cues. In Spanish, adaptation drops sharply for both models, with DeepSeek slightly more responsive than GPT, but both largely retain English-style reasoning. Overall, models adapt their reasoning most in Bengali, moderately in Arabic, and least in Spanish—showing that Linguistic Adaptation reveals how well models can culturally reorient their explanations across languages under the same prompt.

### 6. Discussion

Our findings show that large language models do not simply make random errors when reasoning across languages—they reason differently. These shifts are not signs of fragility, but evidence of cultural adaptability. Each model displays a distinct strategy for handling cultural cues: GPT adapts flexibly to new contexts but often at the cost of consistency; FANAR maintains internal coherence but shows limited adaptability; and DeepSeek strikes a middle ground, flexible yet occasionally erratic.

Statistical analyses support these observations. Kruskal–Wallis tests (all $p < 0.001$, $\varepsilon^2 \approx 0.04$–$0.17$) and Wilcoxon signed-rank tests reveal significant changes in Cultural Fluency across languages. Both GPT ($W = 19611$) and DeepSeek ($W = 11409$) exhibit marked declines from English to Arabic, while showing large gains in Bengali (GPT: $W = 84$; DeepSeek: $W = 8063$). Even in Spanish, where output appeared stable, significant variation persists (GPT: $W = 4331$, $p < 0.001$; DeepSeek: $W = 17822$, $p = 0.00026$). These shifts are not merely lexical or structural—they reflect differences in explanatory framing, such as shifts toward collectivist or pragmatic reasoning styles depending on language and context.

These results highlight a key insight: cultural understanding in LLMs is neither intrinsic nor uniform—it is reconstructed dynamically through language and prompt framing. While models may produce correct answers, their explanations often diverge in tone, emphasis, or moral reasoning. This underscores the need for evaluation methods that probe beyond correctness.

Using CRaFT's four metrics—Cultural Fluency,

Deviation, Consistency, and Linguistic Adaptation—we were able to measure this subtle but critical layer of model behavior. These metrics reveal that what appears as "cultural awareness" is better understood as an emergent, language-contingent property of LLMs, not a stable cognitive trait.

Rather than viewing these inconsistencies as failures, we interpret them as indicators of latent cultural sensitivity—capabilities that are present, but unevenly expressed. Improving this adaptability may be essential for developing AI systems that reason with, rather than over, cultural context.

## 7. Conclusion

We introduced CRaFT, a multilingual, explanation-based evaluation framework for assessing how large language models reason across cultural and linguistic contexts. CRaFT defines four interpretable metrics—Cultural Fluency, Deviation, Consistency, and Linguistic Adaptation—that move beyond correctness to capture variation in explanatory reasoning. Applied to over 2,100 answer–explanation pairs in Arabic, Bengali, and Spanish, these metrics reveal systematic shifts that accuracy-based evaluations fail to detect.

Our findings show that cultural awareness in LLMs is not an intrinsic capability, but an emergent, language-conditioned behavior. Models exhibit distinct trade-offs: some adapt flexibly but inconsistently; others maintain coherence but struggle to generalize culturally. By shifting the focus from what models answer to how they reason, CRaFT offers a new lens for evaluating and improving cultural competence in multilingual language models.

Future work will expand CRaFT to additional languages and cultural contexts, incorporate human evaluations to validate the interpretability of each metric, and explore fine-tuning strategies to enhance cultural reasoning without sacrificing generalization.

## 8. Bibliographical References